%% file: main.tex
\documentclass[10pt,twocolumn,letterpaper]{article}

\usepackage{iccv}
\usepackage{times}
\usepackage{epsfig}
\usepackage{graphicx}
\usepackage{amsmath}
\usepackage{amssymb}
\usepackage[numbers, sort]{natbib}
\usepackage[pagebackref=true,breaklinks=true,letterpaper=true,colorlinks,bookmarks=false]{hyperref}

\usepackage{helvet}   
\usepackage{courier}  
\usepackage{url}      
\usepackage{xcolor}
\usepackage{booktabs}
\usepackage{multirow}
\usepackage{subcaption}
\usepackage{algorithmic}
\usepackage[linesnumbered,ruled,vlined]{algorithm2e}

\usepackage[export]{adjustbox}

\usepackage{color}

\iccvfinalcopy 

\DeclareMathAlphabet      {\mathbfit}{OML}{cmm}{b}{it}

\newcommand{\ra}[1]{\renewcommand{\arraystretch}{#1}}

\newcommand{\xmark}{\ding{55}}
\usepackage{pifont}

\newcommand{\bbR}{{\mathbb{R}}}

\newlength\paramargin
\newlength\figmargin
\newlength\tablemargin
\newlength\secmargin
\newlength\figcapmargin
\newlength\tablecapmargin

\def\iccvPaperID{1182} 

\ificcvfinal\fi
\begin{document}

\title{Cross-Dataset Person Re-Identification\\
via Unsupervised Pose Disentanglement and Adaptation}

\author{Yu-Jhe Li\textsuperscript{1,2,3}, Ci-Siang Lin\textsuperscript{1,2}, Yan-Bo Lin\textsuperscript{1}, Yu-Chiang Frank Wang\textsuperscript{1,2,3}\\
\textsuperscript{1}\hspace{1pt}National Taiwan University, Taiwan\\\textsuperscript{2}\hspace{1pt}MOST Joint Research Center for AI Technology and All Vista Healthcare, Taiwan\\
\textsuperscript{3}\hspace{1pt}ASUS Intelligent Cloud Services, Taiwan\\
{\tt\small \{\url{d08942008}, \url{d08942011}, \url{r06942048}, \url{ycwang}\}\url{@ntu.edu.tw}}
}

\maketitle

\input{0-abstract}

\input{1-intro}

\input{2-related}

\input{3-method}

\input{4-experiment}
\input{5-conclusions}

\vspace{-3mm}
\paragraph{Acknowledgements.} This work is supported by the Ministry of Science and Technology of Taiwan under grant MOST 108-2634-F-002-018.

{
\bibliographystyle{ieee_fullname}
\bibliography{egbib}
}

\end{document}

%% file: 0-abstract.tex
\begin{abstract}

Person re-identification (re-ID) aims at recognizing the same person from images taken across different cameras. On the other hand, cross-dataset/domain re-ID focuses on leveraging labeled image data from source to target domains, while target-domain training data are without label information. In order to introduce discriminative ability and to generalize the re-ID model to the unsupervised target domain, our proposed Pose Disentanglement and Adaptation Network (PDA-Net) learns deep image representation with pose and domain information properly disentangled. Our model allows pose-guided image recovery and translation by observing images from either domain, without pre-defined pose category nor identity supervision. Our qualitative and quantitative results on two benchmark datasets confirm the effectiveness of our approach and its superiority over state-of-the-art cross-dataset re-ID approaches.

\end{abstract}

%% file: 1-intro.tex
\section{Introduction} \label{sec:intro}

Given a query image containing a person (e.g., pedestrian, suspect, etc.), person re-identification (re-ID)~\cite{zheng2016person} aims at matching images with the same identity across non-overlapping camera views. Person re-ID has been among active research topics in computer vision due to its practical applications to smart cities and large-scale surveillance systems. In order to tackle the challenges like visual appearance changes or occlusion in practical re-ID scenarios, several works have been proposed~\cite{ hermans2017defense, lin2017improving,zhong2017camera,si2018dual,chen2018group,shen2018deep}. However, such approaches require a large amount of labeled data for training, and this might not be applicable for real-work applications.


\input{figures/teaser}


Since it might be computationally expensive to collect identity labels for the dataset of interest, one popular solution is to utilize an additional yet distinct source-domain dataset. This dataset contains fully labeled images (but with different identities) captured by a different set of cameras. Thus, the goal of cross-domain/dataset person re-ID is to extract and adapt useful information from source to the target-domain data of interest, so that re-ID at the target-domain can be addressed accordingly. Since no label is observed for the target-domain data during training, one typically views the aforementioned setting as a unsupervised learning task.

Several methods for cross-dataset re-ID have been proposed~\cite{fan2017unsupervised, peng2016unsupervised, yu2017cross, zheng2015scalable, image-image18, wang2018transferable, zhong2018generalizing}. For example, Deng~\etal~\cite{image-image18} employ CycleGAN to covert labeled images from source to target domains, followed by performing re-ID at the target domain. Similarly, Zhong~\textit{et al.}~\cite{zhong2018generalizing} utilize StarGAN~\cite{choi2018stargan} to learn camera invariance and domain connectedness simultaneously. On the other hand, Lin~\etal~\cite{lin2018multi} employ Maximum Mean Discrepancy (MMD) for learning mid-level feature alignment across data domains for cross-dataset re-ID. However, as shown in Fig.~\ref{fig:introduction}, existing cross-domain re-ID approaches generally adapt style information across datasets, and thus pose information cannot be easily be described or preserved in such challenging scenarios.


To overcome the above limitations, we propose a novel deep learning framework for cross-dataset person re-ID. Without observing any ground truth label and pose information in the target domain, our proposed \textit{Pose Disentanglement and Adaptation Network (PDA-Net)} learns domain-invariant features with the ability to disentangle pose information. This allows one to extract, adapt, and manipulate images across datasets without supervision in identity or label. More importantly, this allows us to learn domain and pose-invariant image representation using our proposed network (as depicted in Fig.~\ref{fig:introduction}). With label information observed from the source-domain images for enforcing the re-ID performance, our PDA-Net can be successfully applied to cross-dataset re-ID. Compare to prior unsupervised cross-dataset re-ID approaches which lack the ability to describe pose and content features, our experiments confirm that our model is able to achieve improved performances and thus is practically preferable.

%


We now highlight the contributions of our work below:%
 \begin{itemize}
\item To the best of our knowledge, we are among the first to perform pose-guided yet dataset-invariant deep learning models for cross-domain person re-ID.

\item Without observing label information in the target domain, our proposed PDA-Net learns deep image representation with pose and domain information properly disentangled.
\item The above disentanglement abilities are realized by adapting and recovering source and target-domain images
in a unified framework, simply based on pose information observed from either domain image data.

\item Experimental results on two challenging unsupervised cross-dataset re-ID tasks quantitatively and qualitatively confirm that our method performs favorably against state-of-the-art re-ID approaches.
\end{itemize}

%% file: figures/teaser.tex
\begin{figure}[t!]
	\centering
	\includegraphics[width=0.97\linewidth]{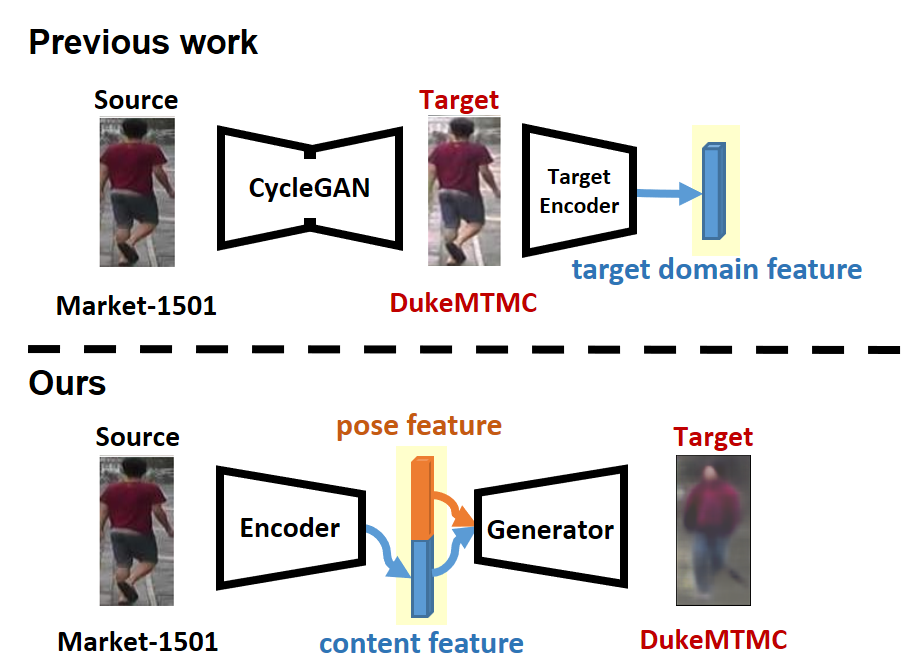}
 	\vspace{-2mm}
     \caption{Existing cross-dataset re-ID methods like~\cite{deng2018image} perform style transfer followed by feature extraction for re-ID, which might limit image variants to be observed. We choose to perform pose disentanglement and adaption with domain-invariant features jointly learned, alleviating the above issue with improved image representation.}
	\vspace{-5mm}
	\label{fig:introduction}
\end{figure}

%% file: 2-related.tex
\section{Related Works} \label{sec:related}


\paragraph{Supervised Person Re-ID.}
Person re-ID has been widely studied in the literature. Existing methods typically focus on tackling the challenges of matching images with viewpoint and pose variations, or those with background clutter or occlusion presented~\cite{li2019greid,li2019crreid,cheng2016person,lin2017improving,kalayeh2018human,si2018dual,chang2018multi,li2018harmonious,liu2018pose,wei2018person,song2018mask,chen2018group,shen2018deep,wang2018resource}. For example, Liu~\etal~\cite{liu2018pose} develop a pose-transferable deep learning framework based on GAN~\cite{goodfellow2014generative} to handle image pose variants. Chen~\etal~\cite{chen2018group} integrate conditional random fields (CRF) and deep neural networks with multi-scale similarity metrics. Several attention-based methods~\cite{si2018dual,chen2018deep,chen2019show,chen2019saliency,lin2018learning,li2018harmonious,song2018mask} are further proposed to focus on learning the discriminative image features to mitigate the effect of background clutter. While promising results have been observed, the above approaches cannot easily be applied for cross-dataset re-ID due to the lack of ability in suppressing the visual differences across datasets.


\vspace{-4mm}
\paragraph{Cross-dataset Person Re-ID.}
To handle cross-dataset person re-ID, a range of hand-crafted features have been considered, so that re-ID at the target domain can be performed in an unsupervised manner~\cite{ma2014covariance,gray2008viewpoint,farenzena2010person,matsukawa2016hierarchical,liao2015person,zheng2015scalable}. To better exploit and adapt visual information across data domains, methods based on domain adaptation~\cite{hoffman2017cycada,chen2019crdoco} have been utilized~\cite{fan2018unsupervised,li2018unsupervised,wang2018transferable,deng2018image,zhong2018generalizing,lin2018multi}. However, since the identities, viewpoints, body poses and background clutter can be very different across datasets, plus no label supervision is available at the target domain, the performance gains might be limited. For example, Fan~\textit{et al.}~\cite{fan2018unsupervised} propose a progressive unsupervised learning method iterating between K-means clustering and CNN fine-tuning. Li~\textit{et al.}~\cite{li2018unsupervised} consider spatial and temporal information to learn tracklet association for re-ID. Wang~\textit{et al.}~\cite{wang2018transferable} learn a discriminative feature representation space with auxiliary attribute annotations. Deng~\textit{et al.}~\cite{deng2018image} translate images from source domain to target domain based on CycleGAN~\cite{CycleGAN2017} to generate labeled data across image domains. Zhong~\textit{et al.}~\cite{zhong2018generalizing} utilize StarGAN~\cite{choi2018stargan} to learn camera invariance features. And, Lin~\textit{et al.}~\cite{lin2018multi} introduce the Maximum Mean Discrepancy (MMD) distance to minimize the distribution variations of two domains.
\input{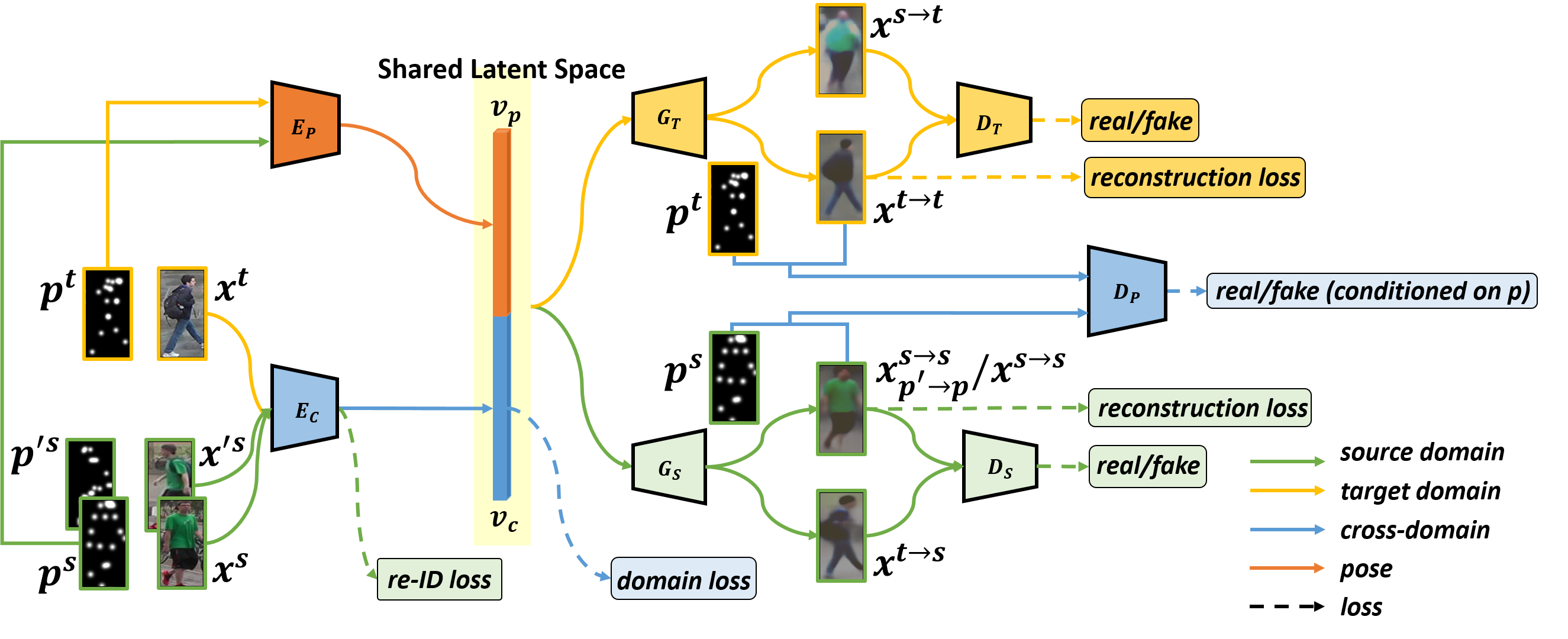}
\vspace{-4mm}
\paragraph{Pose-Guided Re-ID.}
While impressive performances are presented in existing cross-dataset re-ID works, they typically require prior knowledge like the pose of interest, or do not exhibit the ability in describing such information in the resulting features. Recently, a number of models are proposed to better represent pose features during re-ID~\cite{su2017pose, zheng2017pose,zhao2017deeply,zhao2017spindle,li2017learning,yao2019deep,wei2017glad}. Ma~\textit{et al.}~\cite{ma2018disentangled} generate person images by disentangling the input into foreground, background and pose with a complex multi-branch model which is not end-to-end trainable. While Qian~\textit{et al.}~\cite{qian2018pose} are able to generate  pose-normalized images for person re-ID, only eight pre-defined poses can be manipulated. Although Ge~\textit{et al.}~\cite{ge2018fd} learn pose-invariant features with guided image information, their model cannot be applied for cross-dataset re-ID, and thus cannot be applied if the dataset of interest is without any label information. Based on the above observations, we choose to learn dataset and pose-invariant features using a novel and unified model. By disentangling the above representation, re-ID of cross-dataset images can be successfully performed even if no label information is available for target-domain training data.

%% file: figures/model.tex
\begin{figure*}[t!]
	\centering
	\includegraphics[width=1.0\linewidth]{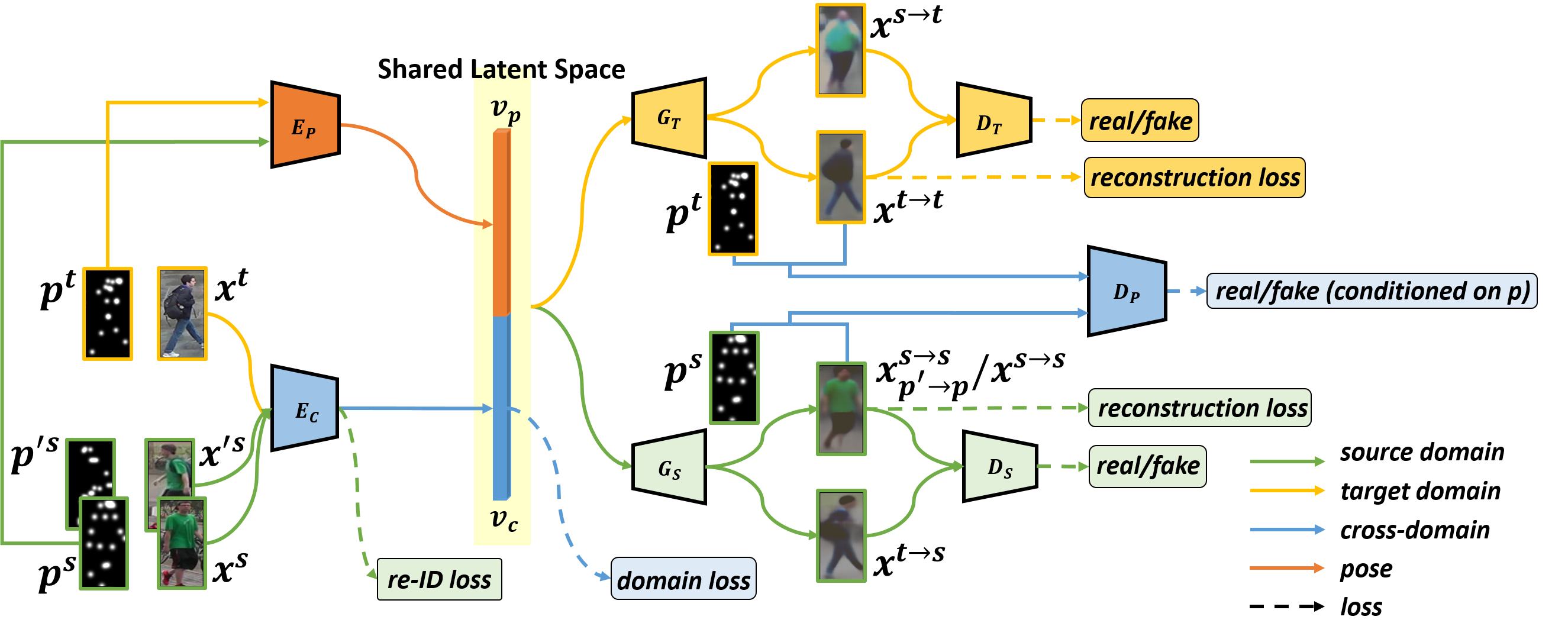}
 	\vspace{-6mm}
    \caption{The overview of our Pose Disentanglement and Adaptation Network (PDA-Net). The content encoder $E_C$ learns domain-invariant features
    $\mathbfit{v}_c$ for input images from either domain. The pose encoder $E_P$ transforms the pose maps ($p^s$ and $p^t$) into the latent features $\mathbfit{v}_p$ for pose guidance and disentanglement purposes. The generators $G_T$ and $G_S$ output domain-specific images via single-domain recovery or cross-domain translation ($x^{s \rightarrow s}_{p^{\prime}\rightarrow p}$, $x^{s \rightarrow s}$, $x^{t \rightarrow s}$, $x^{t \rightarrow t}$ and $x^{s \rightarrow t}$),  conditioned on the pose maps ($p^s$ and $p^t$). The domain discriminators $D_S$ and $D_T$ preserve image perceptual quality, while the pose discriminator $D_P$ is employed for pose disentanglement guarantees.}
	\label{fig:archi}
	\vspace{-5mm}
\end{figure*}

%% file: 3-method.tex
\section{Proposed Method}\label{sec:method}

\subsection{Notations and Problem Formulation}\label{sec:note}

For the sake of completeness, we first define the notations to be used in this paper. Assume that we have the access to a set of $N_S$ images $X_S = \{x_i^s\}_{i=1}^{N_S}$ with the associated label set $Y_S = \{y_i^s\}_{i=1}^{N_S}$, where $x_i^s \in \bbR^{H \times W \times 3}$ and $y_i^s \in \bbR$ represent the $i^{th}$ image in the source-domain dataset and its corresponding identity label, respectively. Another set of $N_T$ target-domain dataset images $X_T = \{x_j^t\}_{j=1}^{N_T}$ without any label information are also available during training, where $x_j^t \in \bbR^{H \times W \times 3}$ represent the $j^{th}$ image in the target-domain dataset. To extract the pose information from source and target-domain data, we apply the pose estimation model~\cite{cao2017realtime} on the above images to generate source/target-domain pose outputs $P_S = \{p_i^s\}_{i=1}^{N_S}$ and $P_T = \{p_j^t\}_{j=1}^{N_T}$, respectively. Note that $p_i^s \in \bbR^{H \times W \times N_L}$ and $p_j^t \in \bbR^{H \times W \times N_L}$ represent the $i^{th}$ and $j^{th}$ pose maps in the corresponding domains, respectively. Following~\cite{cao2017realtime}, we set the number of pose landmarks $N_L=18$ in our work.

To achieve cross-dataset person re-ID, we present an end-to-end trainable network, \emph{Pose Disentanglement and Adaptation Network (PDA-Net)}. As illustrated in Figure~\ref{fig:archi}, our PDA-Net aims at learning domain-invariant deep representation $\mathbfit{v}_c \in \bbR^{d}$ ($d$ denotes the dimension of the feature), while pose information is jointly disentangled from this feature space. To achieve this goal, a pair of encoders $E_C$ and $E_P$ for encoding the input images and pose maps into $\mathbfit{v}_c$ and  $\mathbfit{v}_p \in \bbR^{h}$ ($h$ denotes the dimension of the feature), respectively. Guided by the encoded pose features (from either domain), our domain specific generators ($G_S$ and $G_T$ for source and target-domain datasets, respectively) would recover/synthesize the desirable outputs in the associated data domain. We will detail the properties of each component in the following subsections.

To perform person re-ID of the target-domain dataset in the testing phase, our network encodes the query image by $E_C$ for deriving the domain and pose-invariant representation $\mathbfit{v}_c$, which is applied for matching the gallery ones via nearest neighbor search (in Euclidean distances). 

\subsection{Pose Disentanglement and Adaptation Network (PDA-Net)}\label{ssec:apd-net}

As depicted in Figure~\ref{fig:archi}, our proposed Pose Disentanglement and Adaptation Network consists of a number of network components. The content encoder $E_C$ encodes input images across different domains/datasets and produces content feature $\mathbfit{v}_c$ for person re-ID. The pose encoder $E_P$ encodes the pose maps and produce pose feature $\mathbfit{v}_p$ for pose disentanglement. The two domain-specific generators, $G_S$ and $G_T$, output images in source and target domains respectively (by feeding both $\mathbfit{v}_c$ and $\mathbfit{v}_p$). The two domain specific discriminators, $D_S$ and $D_T$, are designed to enforce the two domain-specific generators $G_S$ and $G_T$ produce perceptually realistic and domain-specific images. Finally, the pose discriminator $D_P$ aims at enforcing the generators to output realistic images conditioned on the given pose.


\subsubsection{Domain-invariant representation for re-ID}


We encourage the content encoder $E_C$ to generate similar feature distributions when observing both $X_S$ and $X_T$. To accomplish this, we apply the Maximum Mean Discrepancy (MMD) measure~\cite{gretton2009fast} to calculate the difference between the associated feature distributions for the content feature $\mathbfit{v}_c$ between the source and target domains. Given an source image $x^s \in X_S$ and an target image $x^t \in X_T$~\footnote{For simplicity, we would omit the subscript $i$ and $j$, denote source and target images as $x^s$ and $x^t$, and represent the corresponding labels for source images as $y^s$ in this paper.}, we first forward $x^s$ and $x^t$ to the content encoder $E_C$ to obtain their content feature $\mathbfit{v}_c^s$ and $\mathbfit{v}_c^t$. Then we can formulate our MMD loss $\mathcal{L}_\mathrm{MMD}$ as:

\begin{equation}\label{eq:mmd}
\begin{aligned}
\mathcal{L}_\mathrm{MMD}  = \| \frac{1}{n_s} \sum_{g = 1}^{n_s} \phi (\mathbfit{v}_{c,g}^{s}) - \frac{1}{n_t} \sum_{l = 1}^{n_t} \phi (\mathbfit{v}_{c,l}^{t}) \|_\mathcal{H}^{2},\\  
\end{aligned}
\end{equation}
where $\phi$ is a map operation which project the distribution into a reproducing kernel Hilbert space $\mathcal{H}$~\cite{gretton2007kernel}. $n_s$ and $n_t$ are the batch sizes of the images in the associated domains. The arbitrary distribution of the features can be represented by using the kernel embedding technique. It has been proven that if the kernel is characteristic, then the mapping to the space $\mathcal{H}$ is injective while the injectivity indicates that the arbitrary probability distribution is uniquely represented by and element in the space $\mathcal{H}$. 

It is also worth noting that, we do \textit{not} consider the adversarial learning strategy for deriving domain-invariant features (e.g., \cite{ganin2016domain}) in our work. This is because that this technique might produce pose-invariant features instead of domain-invariant ones for re-ID datasets, and thus the resulting features cannot perform well in cross-dataset re-ID.


Next, to utilize label information observed from source-domain training data, we impose a triplet loss $\mathcal{L}_{tri}$ on the derived feature vector $\mathbfit{v}_c$. This would maximize the inter-class discrepancy while minimizing intra-class distinctness. To be more specific, for each input source image $x^s$, we sample a positive image $x_\mathrm{pos}^s$ with the same identity label and a negative image $x_\mathrm{neg}^s$ with different identity labels to form a triplet tuple. Then, the distance between $x^s$ and $x_\mathrm{pos}^s$ (or $x_\mathrm{neg}^s$) can be calculated as:
\begin{equation}
  \begin{aligned}
  d_\mathrm{pos} = \|\mathbfit{v}_{c}^s - \mathbfit{v}_{c,\mathrm{pos}}^s\|_2,
  \end{aligned}
  \label{eq:d-pos}
\end{equation}
\begin{equation}
  \begin{aligned}
  d_\mathrm{neg} = \|\mathbfit{v}_{c}^s - \mathbfit{v}_{c,\mathrm{neg}}^s\|_2,
  \end{aligned}
  \label{eq:d-neg}
\end{equation}
where $\mathbfit{v}_{c}^s$, $\mathbfit{v}_{c,\mathrm{pos}}^s$, and $\mathbfit{v}_{c,\mathrm{neg}}^s$ represent the feature vectors of images $x^s$, $x^s_\mathrm{pos}$, and $x^s_\mathrm{neg}$, respectively.

With the above definitions, the triplet loss $\mathcal{L}_{tri}$ is
\begin{equation}
  \begin{aligned}
  \mathcal{L}_{tri}
  = &~ \mathbb{E}_{(x^s,y^s) \sim (X_S,Y_S)}\max(0, m + d_\mathrm{pos} - d_\mathrm{neg}),
  \end{aligned}
  \label{eq:tri}
\end{equation}
where $m > 0$ is the margin enforcing the separation between positive and negative image pairs. 

\subsubsection{Pose-guided cross-domain image translation}

To ensure our derived content feature is  domain-invariant in cross-domain re-ID tasks, we need to perform additional image translation during the learning of our PDA-Net. That is, we have the pose encoder $E_P$ in Fig.~\ref{fig:archi} encodes the inputs from source pose set inputs $P_S$ and the target pose set $P_T$ into pose features $\mathbfit{v}_p^s$ and $\mathbfit{v}_p^t$. As a result, both content and pose features would be produced in the latent space.

We enforce the two generators $G_S$ and $G_T$ for generating the person images conditioned on the encoded pose feature. For the source domain, we have the source generator $G_S$ take the concatenated source-domain content and pose feature pair $(\mathbfit{v}_p^s,\mathbfit{v}_c^s)$ and output the corresponding image ${x}^{s \rightarrow s}$. Similarly, we have $G_T$ take $(\mathbfit{v}_p^t,\mathbfit{v}_c^t)$ for producing ${x}^{t \rightarrow t}$. Note that ${x}^{s \rightarrow s}=G_S((\mathbfit{v}_p^s,\mathbfit{v}_c^s))$, ${x}^{t \rightarrow t}=G_T(\mathbfit{v}_p^t,\mathbfit{v}_c^t)$ denote the reconstructed images in source and target domains, respectively.  Since this can be viewed as image recovery in each domain, reconstruction loss can be applied as the objective during learning.

Since we have ground truth labels (i.e., image pair correspondences) for the source-domain data, we can further perform a unique image recovery task for the source-domain images. To be more precise, given two source-domain images $x^s$ and ${x^{\prime}}^s$ 
of the same person but with different poses $p^s$ and ${p^{\prime}}^s$, we expect that they share the same content feature $\mathbfit{v}_c^s$ but with pose features as $\mathbfit{v}_p^s$ and $\mathbfit{v}_{p^\prime}^s$. Given the desirable pose $\mathbfit{v}_p^s$, we then enforce $G_S$ to output the source domain image $x^s$ using the content feature $\mathbfit{v}_c^s$ which is originally associated with $\mathbfit{v}_{p^\prime}^s$. This is referred to as \textit{pose-guided} image recovery.

With the above discussion, image reconstruction loss for the source-domain data $\mathcal{L}_\mathrm{rec}^S$ can be calculated as:
\begin{equation}
  \label{eq:rec_s}
  \begin{aligned}
  \mathcal{L}_\mathrm{rec}^S = &~ \mathbb{E}_{x^s \sim X_S, p^s \sim P_S}[\|x^{s \rightarrow s} - x^s\|_1]\\
  + &~ \mathbb{E}_{\{x^{s},x^{\prime s}\} \sim X_S, p^s \sim P_S}[\| x_{p^{\prime} \rightarrow p}^{s \rightarrow s} - x^{s}\|_1],
  \end{aligned}
\end{equation}
where $x_{p^{\prime} \rightarrow p}^{s \rightarrow s} = G_S(\mathbfit{v}_{p}^s,\mathbfit{v}_c^s| \mathbfit{v}_{p^\prime}^s)$ denotes the generated image from the input $x^{\prime s}$ and $v^s_c$ describe the content feature of the same identity (i.e., $x^{\prime s}$, and $x^s$ of the same person by with different poses $p^{\prime}$ and $p$).

\noindent As for the target-domain reconstruction loss, we have 
\begin{equation}
  \label{eq:rec_t}
  \begin{aligned}
  \mathcal{L}_\mathrm{rec}^T = &~ \mathbb{E}_{x^t \sim X_T, p^t \sim P_T}[\|x^{t \rightarrow t} - x^t\|_1].
  \end{aligned}
\end{equation}
Note that we adopt the L$1$ norm in the above reconstruction loss terms as it preserves image sharpness~\cite{huang2018munit}. 

In addition to image recovery in either domain, our model also perform pose-guided image translation. That is, our decoders $G_S$ and $G_T$ allow input feature pairs whose content and pose representation are extracted from different domains. Thus, we would observe ${x}^{t \rightarrow s}=G_S(\mathbfit{v}_p^t,\mathbfit{v}_c^t)$ and ${x}^{s \rightarrow t}=G_T(\mathbfit{v}_p^s,\mathbfit{v}_c^s)$ as the outputs, with the goal of having these translated images as realistic as possible.

To ensure $G_S$ and $G_T$ produce perceptually realistic outputs in the associated domains, we have the image discriminator $D_S$ discriminate between the real source-domain images ${x}^s$ and the synthesized/translated ones (i.e., ${x}^{s \rightarrow s}$, ${x}^{t \rightarrow s}$). Thus, the source-domain discriminator loss $\mathcal{L}_{domain}^S$ as
\begin{equation}
  \begin{aligned}
  \mathcal{L}_{domain}^S = &~ \mathbb{E}_{x^s \sim X_S}[\log(\mathcal{D}_{S}(x^s))] \\
  + & ~ \mathbb{E}_{x^s \sim X_S, p^s \sim P_S}[\log(1 - \mathcal{D}_{S}({x}^{s \rightarrow s}))] \\
  %
  %
  + & ~ \mathbb{E}_{x^t \sim X_T, p^t \sim P_T}[\log(1 - \mathcal{D}_{S}({x}^{t \rightarrow s}))].
  \end{aligned}
  \label{eq:src_style_loss}
\end{equation}
\vspace{-1mm}
\noindent Similarly, the target domain discriminator loss $\mathcal{L}_{domain}^T$ is defined as
%
\begin{equation}
  \begin{aligned}
  \mathcal{L}_{domain}^T = &~ \mathbb{E}_{x^t \sim X_T}[\log(\mathcal{D}_{T}(x^t))] \\
  + & ~ \mathbb{E}_{x^t \sim X_T, p^t \sim P_T}[\log(1 - \mathcal{D}_{T}({x}^{t \rightarrow t}))] \\
  + & ~ \mathbb{E}_{x^s \sim X_S, p^s \sim P_S}[\log(1 - \mathcal{D}_{T}({x}^{s \rightarrow t}))].
  \end{aligned}
  \label{eq:tgt_style_loss}
\end{equation}

\subsubsection{Unsupervised pose disentanglement across data domains}

With the above pose-guided image translation mechanism, we have our PDA-Net learn domain-invariant content features across data domains. However, to further ensure the pose encoder describes and disentangles the pose information observed from the input images, we need additional network modules for completing this goal.

To achieve this object, we introduce a pose discriminator $D_P$ in Fig.~\ref{fig:archi}, which focuses on distinguishing between real and recovered images, conditioned on the given pose inputs. Following previous FD-GAN~\cite{ge2018fd}, we adopt the PatchGAN~\cite{pix2pix} structure as our $D_P$. That is, the input to $D_P$ is concatenation of the real/recovered image and the given pose map, which is processed by Gaussian-like heat-map transformation. Then, $D_P$ produces a image-pose matching confidence map, each location of this output confidence map represents the matching degree between the input image and the associated pose map. 
\input{psuedo}
It can be seen that, the two generators $G_S$ and $G_T$ in PDA-Net tend to fool the pose discriminator $D_P$ to obtain high matching confidences for the generated images. Intuitively, since only source-domain data are with ground truth labels, our $D_P$ is designed to authenticate the recovered images in each corresponding domain but not the translated ones across domains. In other words, the adversarial loss of $D_P$ is formulated as:
\begin{equation}
  \begin{aligned}
  \mathcal{L}_{pose} = \mathcal{L}_{pose}^S + \mathcal{L}_{pose}^T,
  \end{aligned}
  \label{eq:pose_loss}
\end{equation}
where
\begin{equation}
  \begin{aligned}
  \mathcal{L}_{pose}^S = &~ \mathbb{E}_{x^s \sim X_S, p^s \sim P_S}[\log(\mathcal{D}_{P}(p^s,x^s))] \\
  + & ~ \mathbb{E}_{x^s \sim X_S, p^s \sim P_S}[\log(1 - \mathcal{D}_{P}(p^s,{x}^{s \rightarrow s}))] \\
  + & ~ \mathbb{E}_{x^s \sim X_S, p^{\prime s} \sim P_S}[\log(1 - \mathcal{D}_{P}(p^{\prime s},{x}^s))] \\
  %
  %
  + & ~ \mathbb{E}_{\{x^s,x^{\prime s}\} \sim X_S, p^{s} \sim P_S}[\log(1 - \mathcal{D}_{P}(p^s,x_{p^{\prime} \rightarrow p}^{s \rightarrow s}))] \\
  %
  %
  \end{aligned}
  \label{eq:src_pose_loss}
\end{equation}
and
\begin{equation}
  \begin{aligned}
  \mathcal{L}_{pose}^T = &~ \mathbb{E}_{x^t \sim X_T, p^t \sim P_T}[\log(\mathcal{D}_{P}(p^t,x^t))] \\
  + & ~ \mathbb{E}_{x^t \sim X_T, p^t \sim P_T}[\log(1 - \mathcal{D}_{P}(p^t,x^{t \rightarrow t}))]. \\
  %
  \end{aligned}
  \label{eq:tgt_pose_loss}
\end{equation}
Note that $x_{p^{\prime} \rightarrow p}^{s \rightarrow s} = G_S(\mathbfit{v}_{p}^s,\mathbfit{v}_c^s| \mathbfit{v}_{p^\prime}^s)$ represents the synthesized image from the input $x^{\prime s}$ (with the same content feature $v_c^s$ with $x^s$ but with a different pose feature $v_p^{\prime s}$). 

From~\eqref{eq:pose_loss}, we see that while our pose disentanglement loss enforces the matching between the output image and its conditioned pose in each domain, additional guidance is available in the source domain to update our $D_P$. That is, as shown in~\eqref{eq:src_style_loss}, we are able to verify the authenticity of the source-domain output image which is given by the input image of the same person but with a different pose (i.e., $p^\prime$ instead of $p$). While our decoder is able to output such a image with its ground truth source-domain image observed (as noted in~\eqref{eq:rec_s}, the introduced $D_P$ would further improve our capability of pose disentanglement and pose-guided image recovery.

It is worth repeating that the goal of PDA-Net is to perform cross-dataset re-ID without observing label information in the target domain. By introducing the aforementioned network module, our PDA-Net would be capable of performing cross-dataset re-ID via pose-guided cross-domain image translation. More precisely, with the joint training of cross-domain encoders/decoders and the pose disentanglement discriminators, our model allows learning of domain-invariant and pose-disentangled feature representation. The pseudo code for training our PDA-Net is summarized in Algorithm \ref{alg:pdanet}.

%% file: psuedo.tex
\begin{algorithm}[t]
\small
\KwData{Source domain: $X_{S}$, $P_S$, and $Y_{S}$; Target domain: $X_{T}$ and $P_T$
}
\KwResult{Configurations of PDA-Net}
$\theta_{E_C}$, $\theta_{E_P}$, $\theta_{G_S}$, $\theta_{G_T}$, $\theta_{D_S}$, $\theta_{D_T}$, $\theta_{D_P}$ $\leftarrow$ initialize\\
  \For{Num. of training Iters.}{
    $x^s$, $p^s$, $y^s$, $x^t$, $p^t$, $x^{\prime s}$, $p^{\prime s}$  $\leftarrow$ sample from $X_S$, $P_S$, $Y_S$, $X_T$, $P_T$\\
    $v_{c}^s$, $v_{c}^t$ $\leftarrow$ obtain by  $E_C(x^s/x^{\prime s})$, $E_C(x^t)$\\
    $v_{p}^s$, $v_{p}^t$ $\leftarrow$ obtain by $E_P(p^s)$, $E_P(p^t)$\\
    $\mathcal{L}_\mathrm{MMD}$, $\mathcal{L}_{tri} \leftarrow$ calculate by (\ref{eq:mmd}), (\ref{eq:tri})\\
    $\theta_{E_C} \xleftarrow{+} -{\nabla}_{\theta_{E_C}}(\mathcal{L}_\mathrm{MMD}+ \lambda_{tri} \mathcal{L}_{tri})$\\
    
    ${x}^{s \rightarrow s}$, ${x}^{t \rightarrow s}$ $\leftarrow$ obtain by $G_S(\mathbfit{v}_p^s,\mathbfit{v}_c^s)$, $G_S(\mathbfit{v}_p^t,\mathbfit{v}_c^t)$\\
    
    ${x}^{s \rightarrow t}$, ${x}^{t \rightarrow t}$ $\leftarrow$ obtain by $G_T(\mathbfit{v}_p^s,\mathbfit{v}_c^s)$, $G_T(\mathbfit{v}_p^t,\mathbfit{v}_c^t)$\\
    $x_{p^{\prime} \rightarrow p}^{s \rightarrow s}$  $\leftarrow$ obtain by $G_S(\mathbfit{v}_{p}^s,\mathbfit{v}_c^s| \mathbfit{v}_{p^\prime}^s)$\\
    %
    $\mathcal{L}_{rec}^S$, $\mathcal{L}_{rec}^T$, $\mathcal{L}_{domain}^S$, $\mathcal{L}_{domain}^T$,  $\mathcal{L}_{pose}$ $\leftarrow$ calculate by  (\ref{eq:rec_s}), (\ref{eq:rec_t}), (\ref{eq:src_style_loss}), (\ref{eq:tgt_style_loss}), (\ref{eq:pose_loss})\\
    %
    \For{  Iters. of updating generator }{
    $\theta_{E_C, E_P, G_S} \xleftarrow{+} -{\nabla}_{\theta_{E_C, E_P, G_S}}(\lambda_{rec}\mathcal{L}_{rec}^S - \mathcal{L}_{domain}^S- \lambda_{pose}\mathcal{L}_{pose})$\\
    $\theta_{E_C, E_P, G_T} \xleftarrow{+} -{\nabla}_{\theta_{E_C, E_P, G_T}}(\lambda_{rec}\mathcal{L}_{rec}^T - \mathcal{L}_{domain}^T- \lambda_{pose}\mathcal{L}_{pose})$\\
  	}
    \For{Iters. of updating discriminator }{
    $\theta_{D_S} \xleftarrow{+} -{\nabla}_{\theta_{D_S}}\mathcal{L}_{domain}^S$\\
    $\theta_{D_T} \xleftarrow{+} -{\nabla}_{\theta_{D_T}}\mathcal{L}_{domain}^T$\\
    $\theta_{D_P} \xleftarrow{+} -{\nabla}_{\theta_{D_P}}\mathcal{L}_{pose}$\\
  	}
  }
\caption{Learning of PDA-Net}\label{alg:pdanet}
\normalsize
\end{algorithm}


%% file: 4-experiment.tex
\section{Experiments} \label{sec:results}

\input{exp/market}

\subsection{Datasets and Experimental Settings}\label{sec:dataset}

To evaluate our proposed method, we conduct experiments on Market-1501~\cite{zheng2015scalable} and DukeMTMC-reID~\cite{zheng2017unlabeled,ristani2016MTMC}, both are commonly considerd in recent re-ID tasks.  
\vspace{-5mm}
\paragraph{Market-1501.} The Market-1501~\cite{zheng2015scalable} is composed of  32,668 labeled images of 1,501 identities collected from 6 camera views. The dataset is split into two non-over-lapping fixed parts: 12,936 images from 751 identities for training and 19,732 images from 750 identities for testing. In testing, 3368 query images from 750 identities are used to retrieve the matching persons in the gallery. 

\vspace{-5mm}
\paragraph{DukeMTMC-reID.} The DukeMTMC-reID~\cite{zheng2017unlabeled, ristani2016MTMC} is also a large-scale Re-ID dataset. It is collected from 8 cameras and contains 36,411 labeled images belonging to 1,404 identities. It also consists of 16,522 training images from 702 identities, 2,228 query images from the other 702 identities, and 17,661 gallery images. 
\vspace{-3mm}
\paragraph{Evaluation Protocol.} We employ the standard metrics as in most person Re-ID literature, namely the cumulative matching curve (CMC) used for generating ranking accuracy, and the mean Average Precision (mAP). We report rank-1 accuracy and mean average precision (mAP) for evaluation on both datasets. 

\input{exp/duke.tex}
\input{exp/ablation}

\subsection{Implementation Details}

\paragraph{Configuration of PDA-Net.}

We implement our model using PyTorch. Following Section~\ref{sec:method}, we use ResNet-$50$ pre-trained on ImageNet as our backbone of cross-domain encoder $E_C$. Given an input image $x$ (all images are resized to size $256 \times 128 \times 3$, denoting width, height, and channel respectively.), $E_C$ encodes the input into $2048$-dimension content feature $\mathbfit{v}_c$. As mentioned in the Section.~\ref{sec:note}, the pose-map is represented by an $18$-channel map, where each channel represents the location of one pose landmark. Such landmark location is converted to a Gaussian heat map. The pose encoder $E_P$ then employs $4$ convolution blocks to produce the $256$-dimension pose feature vector $\mathbfit{v}_p$ from these pose-maps. The structure of the both the domain generators ($G_S$, $G_T$) are $6$ convolution-residual blocks similar to that proposed by Miyato~\etal~\cite{miyato2018cgans}. The structure of the both the domain discriminator (${D}_S$, ${D}_T$) employ the ResNet-$18$ as backbone while the architecture of shared pose disciminator $D_P$ adopts PatchGAN structure following FD-GAN~\cite{ge2018fd} and is composed of $5$ convolution blocks in our PDA-Net. Domain generators ($G_S$, $G_T$), domain discriminator (${D}_S$, ${D}_T$), shared pose discriminator $D_P$ are all randomly initialized. The margin for the $\mathcal{L}_{tri}$ is set as $0.5$, and we fix $\lambda_{tri}$, $\lambda_{rec}$, and $\lambda_{pose}$ as $1.0$, $10.0$, $0.1$, respectively.

\subsection{Quantitative Comparisons}\label{sec:sta}
\vspace{-2mm}
\paragraph{Market-1501.} In Table~\ref{tab:market}, we compare our proposed model with the use of Bag-of-Words (BoW)~\cite{zheng2015scalable} for matching (i.e., no transfer), four unsupervised re-ID approaches, including UMDL~\cite{peng2016unsupervised}, PUL~\cite{fan2017unsupervised}, CAMEL~\cite{yu2017cross} and TAUDL~\cite{li2018unsupervised}, and seven cross-dataset re-ID methods, including PTGAN~\cite{wei2018person}, SPGAN~\cite{deng2018image}, TJ-AIDL~\cite{wang2018transferable}, MMFA~\cite{lin2018multi}, HHL~\cite{zhong2018generalizing}, CFSM~\cite{chang2018disjoint} and ARN~\cite{Li_2018_CVPR_Workshops}. From this table, we see that our model achieved very promising results in Rank-1, Rank-5, Rank-10, and mAP, and observed performance margins over recent approaches. For example, in the single query setting, we achieved~\textbf{Rank-1 accuracy=75.2\%} and~\textbf{mAP=52.6\%}.

Compared to SPGAN~\cite{deng2018image} and HHL~\cite{zhong2018generalizing}, we note that our model is able to generate cross-domain images conditioned on various poses rather than few camera styles. Compared to MMFA~\cite{lin2018multi}, our model further disentangles the pose information and learns a pose invariant cross-domain latent space. Compared to the second best method,~\textit{i.e.}, TAUDL~\cite{li2018unsupervised}, our results were higher by~\textbf{11.5\%} in Rank-1 accuracy and by~\textbf{11.4\%} in mAP, while no additional spatial and temporal information is utilized (but TAUDL did).
\vspace{-5mm}

\paragraph{DukeMTMC-reID.} We now consider the DukeMTMC-reID as the target-domain dataset of interest, and list the comparisons in Table~\ref{tab:duke}. From this table, we also see that our model performed favorably against baseline and state-of-art  unsupervised/cross-domain re-ID methods. Take the single query setting for example, we achieved~\textbf{Rank-1 accuracy=63.2\%} and~\textbf{mAP=45.1\%}. Compared to the second best method, our results were higher by~\textbf{1.5\%} in Rank-1 accuracy and by~\textbf{1.6\%} in mAP. From the experiments on the above two datasets, the effectiveness of our model for cross-domain re-ID can be successfully verified.

\input{exp/visual}
\subsection{Ablation Studies and Visualization}\label{sec:abl}
\paragraph{Analyzing the network modules in PDA-Net.}
As shown in Table~\ref{table:exp-abla}, we start from two baseline methods, i.e., naive Resnet-$50$ (w/o $\mathcal{L}_\mathrm{MMD}$) and advanced Resnet-$50$ (w/ $\mathcal{L}_\mathrm{MMD}$), showing the standard re-ID performances. We then utilize ResNet-$50$ as the backbone CNN model to derive representations for re-ID with only triplet loss $\mathcal{L}_{tri}$, while the advanced one includes the MMD loss $\mathcal{L}_\mathrm{MMD}$. We observe that our full model (the last row) improved the performance by a large margin (roughly $20\sim25\%$) at Rank-1 on both two benchmark datasets. The performance gain can be ascribed to the unique design of our model for deriving both domain-invariant and pose-invariant representation.
\vspace{-3mm}
\paragraph{Loss functions}
To further analyze the importance of each introduced loss function, we conduct an ablation study from third row to seventh rows shown in Table~\ref{table:exp-abla}. Firstly, the reconstruction loss $\mathcal{L}_{rec}$ is shown to be vital to our PDA-Net, since we observe $23\%$ and $20\%$ drops on Market-1501 and DukeMTMC-reID, respectively when the loss was excluded. This is caused by no explicit supervision to guide our PDA-Net to generate human-perceivable images, and thus the resulting model would suffer from image-level information loss.

Secondly, without the pose loss $\mathcal{L}_{pose}$ on both domains, our model would not be able to perform pose matching based on each generated image, causing failure on the pose disentanglement process and resulting in the re-ID performance drop (about $20\%$ on both settings). Thirdly, when $\mathcal{L}_{domain}^{S/T}$ is turned off, our model is not able to preserve the domain information, indicating that only pose information would be observed. We credited such a $10\%$ performance drop to the negative effect in learning pose-invariant feature, which resulted in unsatisfactory pose disentanglement. Lastly, the MMD loss $\mathcal{L}_\mathrm{MMD}$ is introduced to our PDA-NET to mitigate the domain shift due to dataset differences. Its effectiveness is also confirmed by our studies.

\input{figures/exp}
\paragraph{Shared pose discriminator $D_P$.}
To demonstrate the effectiveness and necessity of the pose discriminator $D_P$ introduced to our PDA-Net, we first consider replacing $D_P$ by two separate pose discriminators $D_P^S$ and $D_P^T$, and report the re-ID performance in the fifth row of Table~\ref{table:exp-abla}. With a clear performance drop observed, we see that the resulting PDA-Net would not be able to transfer the substantiated pose-matching knowledge from source to target domains. In other words, a shared pose discriminator would be preferable since pose guidance can be provided by both domains. 

\paragraph{Visualization comparisons of cross-dataset and pose-guided re-ID models.}
In Figure~\ref{fig:exp-vis}, we visualize the generated images: ${x}^{s \rightarrow s}$,  ${x}^{s \rightarrow t}$,  ${x}^{t \rightarrow s}$, and  ${x}^{t \rightarrow t}$ in two cross-domain settings. Given an input from either domain with pose conditions, our model was able to produce satisfactory pose-guided image synthesis within or across data domains. 

In Figure~\ref{fig:exp}, we additionally consider the cross-dataset re-ID appoach of SPGAN~\cite{image-image18} and the pose-disentanglement re-ID method of FD-GAN~\cite{ge2018fd}.  We see that, since SPGAN performed style transfer for synthesizing cross-domain images, pose variants cannot be exploited in the target domain. While FD-GAN was able to generate pose-guided image outputs with supervision on target target domain, their model is not designed to handle cross-domain data so that cannot produce images across datasets with satisfactory quality. From the above qualitative evaluation and comparison, we confirm that our PDA-Net is able to perform pose-guided single-domain image recovery and cross-domain image translation with satisfactory image quality, which would be beneficial to cross-domain re-ID tasks.
\vspace{-3mm}

%% file: exp/market.tex
\begin{table}[t]
\centering
\vspace{-2mm}
\caption{Performance comparisons on Market-1501 with cross-dataset/unsupervised Re-ID methods. The number in bold indicates the best result.}
\label{tab:market}
\resizebox{\linewidth}{!} 
  {
\begin{tabular}{ l|c c c c }
\toprule
\multirow{2}{*}{Method} & \multicolumn{4}{c}{\begin{tabular}[c]{@{}l@{}}Source: DukeMTMC, Target: Market \end{tabular}} \\
& Rank-1 & Rank-5 & Rank-10 & mAP         \\
\midrule
BOW~\cite{zheng2015scalable}  & 35.8   & 52.4   & 60.3    & 14.8  \\ 
UMDL~\cite{peng2016unsupervised}  & 34.5   & 52.6   & 59.6    & 12.4  \\
PTGAN~\cite{wei2018person}      & 38.6  & -   & 66.1    & -   \\
PUL~\cite{fan2017unsupervised}     & 45.5   & 60.7   & 66.7    & 20.5  \\
CAMEL~\cite{yu2017cross}     & 54.5   & -  & -   & 26.3    \\ 
SPGAN~\cite{image-image18}    & 57.7   & 75.8   & 82.4    & 26.7   \\
TJ-AIDL~\cite{wang2018transferable}      & 58.2   & 74.8   & 81.1   & 26.5   \\
MMFA~\cite{lin2018multi}      & 56.7   & 75.0   & 81.8    & 27.4   \\
HHL~\cite{zhong2018generalizing}      & 62.2   & 78.8   & 84.0    & 31.4   \\
CFSM~\cite{chang2018disjoint}      & 61.2   & -   & -    & 28.3   \\
ARN~\cite{Li_2018_CVPR_Workshops}	& 70.3 & 80.4 & 86.3 &39.4\\ 
TAUDL~\cite{li2018unsupervised}      & 63.7   & -   & -    & 41.2   \\
\textbf{PDA-Net (Ours)}  &   \textbf{75.2} &  \textbf{86.3} & \textbf{90.2}  &  \textbf{47.6} \\ 
\bottomrule
\end{tabular}
}
\vspace{-2mm}
\end{table}


%% file: exp/duke.tex
\begin{table}[t]
\centering
\vspace{-2mm}
\caption{Performance comparisons on DukeMTMC-reID with cross-dataset/unsupervised Re-ID methods. The number in bold indicates the best result.}
\label{tab:duke}
\resizebox{\linewidth}{!} 
  {
\begin{tabular}{ l|c c c c }
\toprule
\multirow{2}{*}{Method} & \multicolumn{4}{c}{\begin{tabular}[c]{@{}l@{}}Source: Market, Target: DukeMTMC\end{tabular}} \\
 & Rank-1 & Rank-5 & Rank-10 & mAP         \\
\midrule
BOW~\cite{zheng2015scalable}       & 17.1  & 28.8   & 34.9    & 8.3\\ 
UMDL~\cite{peng2016unsupervised}      & 18.5   & 31.4   & 37.6    & 7.3  \\ 
PTGAN~\cite{wei2018person}      & 27.4  & -   & 50.7    & -   \\
PUL~\cite{fan2017unsupervised}     & 30.0     & 43.4   & 48.5    & 16.4 \\
SPGAN~\cite{image-image18}     & 46.4   & 62.3   & 68.0      & 26.2 \\ 
TJ-AIDL~\cite{wang2018transferable}      & 44.3   & 59.6   & 65.0   & 23.0   \\
MMFA~\cite{lin2018multi}      & 45.3   & 59.8   & 66.3    & 24.7   \\
HHL~\cite{zhong2018generalizing}      & 46.9   & 61.0   & 66.7    & 27.2   \\
CFSM~\cite{chang2018disjoint}      & 49.8   & -   & -    & 27.3   \\
ARN~\cite{Li_2018_CVPR_Workshops}	& 60.2   & 73.9  & 79.5   &33.4   \\ 
TAUDL~\cite{li2018unsupervised}      & 61.7   & -   & -    & 43.5   \\
\textbf{PDA-Net (Ours)}  &   \textbf{63.2} &  \textbf{77.0} & \textbf{82.5}  &  \textbf{45.1} \\ 
\bottomrule
\end{tabular}
}
\vspace{-2mm}
\end{table}

%% file: exp/ablation.tex
\begin{table*}[!tbp]
  \small
  \ra{1.3}
  \begin{center}
  \caption{
Ablation studies of the proposed PDA-Net under two experimental settings. ``Share $D_P$" incidates whether to build separate pose discriminators, i.e. $D_P^S$ and $D_P^T$, instead of one shared $D_P$. 
  }
  \vspace{\tablecapmargin}
  \label{table:exp-abla}
  \resizebox{1\linewidth}{!} 
  {
  \begin{tabular}{l|cccccc|cc|cc}
  \toprule
  \multirow{2}{*}{Experimental setting} & \multicolumn{6}{c|}{Loss functions and component} & \multicolumn{2}{c|}{\begin{tabular}[c]{@{}l@{}}Source: DukeMTMC-reID\\Target: Market-1501 \end{tabular}} & \multicolumn{2}{c}{\begin{tabular}[c]{@{}l@{}}Source: Market-1501\\ Target: DukeMTMC-reID\end{tabular}} \\
  \cmidrule{2-11}
  &  $\mathcal{L}_{tri}$ & $\mathcal{L}_\mathrm{MMD}$ & $\mathcal{L}_{rec}^{S/T}$ & $\mathcal{L}_{domain}^{S/T}$ &  $\mathcal{L}_{pose}$ & Share $D_P$  & Rank-1 & mAP & Rank-1 & mAP\\
  \midrule
  Baseline (ResNet-$50$) & \checkmark & \xmark & \xmark & \xmark & \xmark & \xmark & 44.2 & 18.1 & 33.5 & 16.3 \\
  Baseline (ResNet-$50$ w/ MMD ) & \checkmark & \checkmark & \xmark & \xmark & \xmark & \xmark & 50.4 & 22.6 & 39.5 & 23.1 \\
  %
  %
  PDA-Net (w/o $\mathcal{L}_{rec}^S$,$\mathcal{L}_{rec}^T$) & \checkmark & \checkmark & \xmark &  \checkmark & \checkmark & \checkmark & 52.3 & 24.7 & 42.5 & 24.0 \\
  PDA-Net (w/o $\mathcal{L}_{pose}$) & \checkmark & \checkmark & \checkmark &  \checkmark & \xmark & \checkmark & 55.1 & 25.2 & 45.5 & 26.1 \\
  PDA-Net (w/o share $D_P$) & \checkmark & \checkmark & \checkmark &  \checkmark & \checkmark & \xmark  & 59.4 & 27.8 & 50.9 & 29.7 \\
  %
  %
  PDA-Net (w/o $\mathcal{L}_{domain}^S, \mathcal{L}_{domain}^T$) & \checkmark & \checkmark & \checkmark &  \xmark & \checkmark & \checkmark  & 65.3 & 30.7 & 56.5 & 31.2 \\
  PDA-Net (w/o MMD) & \checkmark & \xmark & \checkmark & \checkmark & \checkmark & \checkmark & 71.2 & 39.8 & 60.1 & 35.8 \\
  
  PDA-Net (Ours) & \checkmark & \checkmark & \checkmark & \checkmark & \checkmark & \checkmark & \textbf{75.2} &  \textbf{47.6} &  \textbf{63.2} &  \textbf{45.1} \\
  \bottomrule
  \end{tabular}
  }
  \end{center}
  \vspace{-4.0mm}
\end{table*}

%% file: exp/visual.tex
\begin{figure*}[t]
  \centering
  \includegraphics[width=0.95\linewidth]{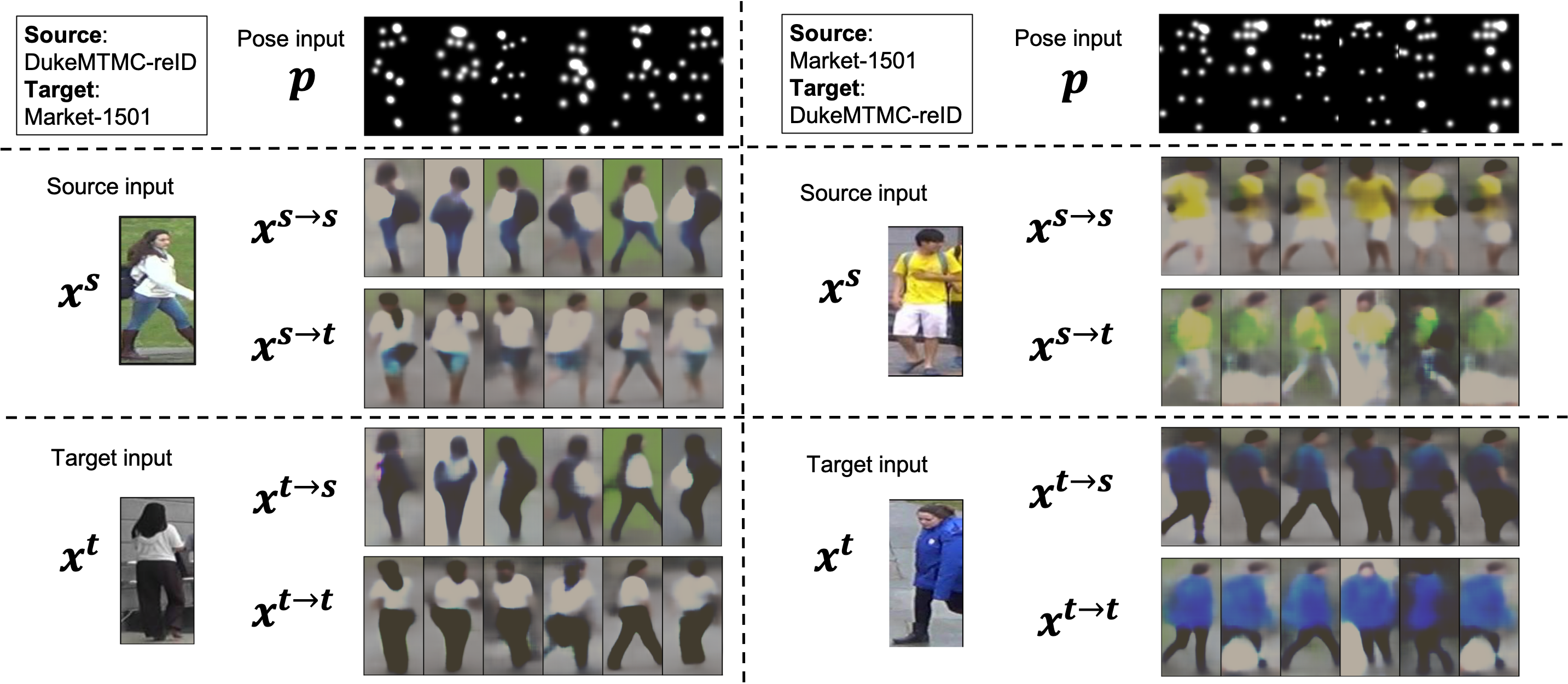}
   \vspace{-1.0mm}
  \caption{Visualization examples of our PDA-Net for pose-guided image translation across datasets. Given six pose conditions (the first row) and the input image ($x^s$ or $x^t$), we present the six generated images for each dataset pair: $x^{s \rightarrow s}$ (the second row), $x^{t \rightarrow s}$ (the third row), $x^{t \rightarrow t}$ (the fourth row) and $x^{s \rightarrow t}$ (the fifth row).}
  \label{fig:exp-vis}
  \vspace{-5mm}
\end{figure*}

%% file: figures/exp.tex
\begin{figure}[t!]
	\centering
	\includegraphics[width=1.0\linewidth]{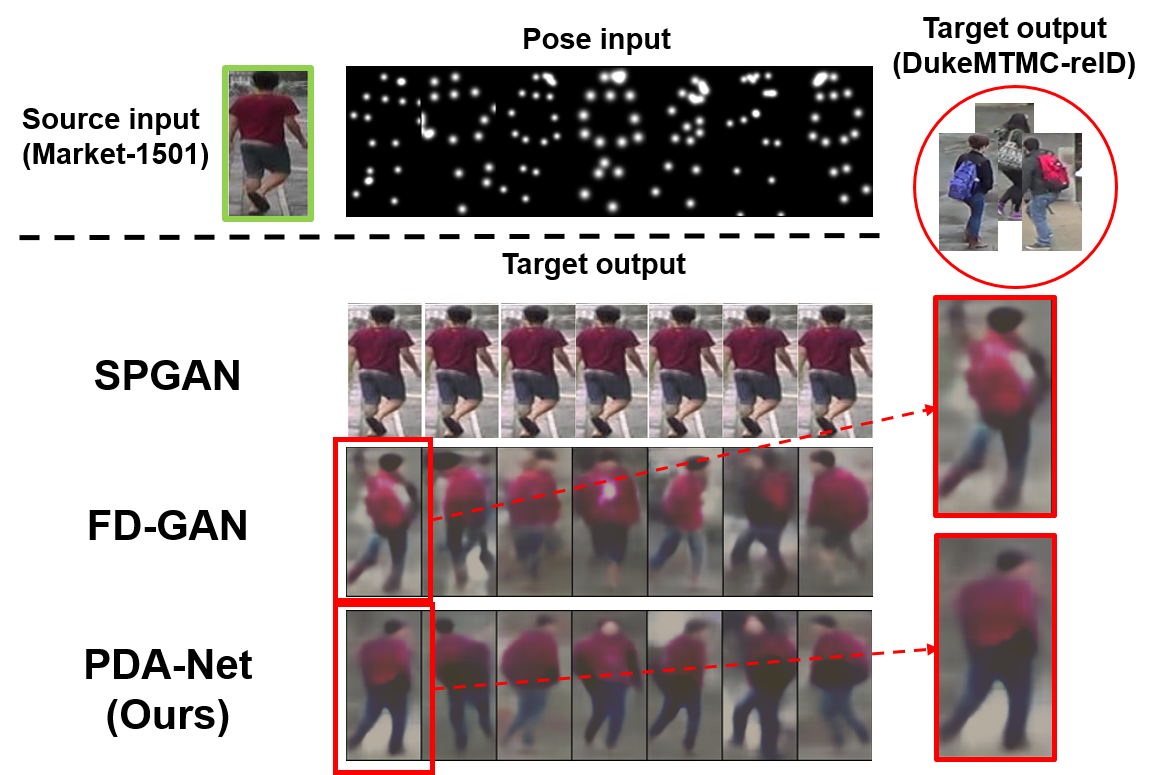}
 	\vspace{-4mm}
     \caption{Visualization of cross-dataset or pose-guided re-ID. Note that SPGAN~\cite{image-image18} performs style-transfer for converting images across datasets but lacks the ability to exhibit pose variants, while FD-GAN~\cite{ge2018fd} disentangles pose information but cannot take cross-domain data.}
	\vspace{-5mm}
	\label{fig:exp}
\end{figure}

%% file: 5-conclusions.tex
\section{Conclusions} \label{sec:conclusions}
In this paper, we presented a novel Pose Disentanglement and Adaptation Network (PDA-Net) for cross-dataset re-ID. The main novelty lies in the unique design of our PDA-Net, which jointly learns domain-invariant and pose-disentangled visual representation with re-ID guarantees. By observing only image input (from either domain) and any desirable pose information, our model allows pose-guided singe-domain image recovery and cross-domain image translation. Note that only label information (image correspondence pairs) is available for the source-domain data, any no pre-defined pose category is utilized during training. Experimental results on the two benchmark datasets showed remarkable improvements over existing works, which support the use of our proposed approach for cross-dataset re-ID. Qualitative results also confirmed that our model is capable of performing cross-domain image translation with pose properly disentangled/manipulated.


